\documentclass{article}

\RequirePackage{fix-cm}
\usepackage{graphicx}
\usepackage{subfigure}
\usepackage [accepted]{icml2019}
\usepackage{helvet}  
\usepackage{courier}  
\usepackage{url}  
\usepackage{times}
\usepackage{xcolor}
\usepackage{soul}
\usepackage[small]{caption} 
\usepackage{algorithm}
\usepackage{algorithmic}
\usepackage{amssymb}
\usepackage{amsmath}
\usepackage{dsfont}
\usepackage{breakcites}
\usepackage{stmaryrd}
\usepackage{lineno}
\usepackage{amsthm}
\usepackage{natbib}
\usepackage[utf8]{inputenc} 
\usepackage[T1]{fontenc}    
\usepackage{booktabs}       
\usepackage{amsfonts}       
\usepackage{nicefrac}       
\usepackage{microtype}      
\usepackage{xargs}       

\usepackage[colorinlistoftodos,prependcaption,textsize=tiny]{todonotes}

\newtheorem{theorem}{Theorem}

\newtheorem{corollary}{Corollary}

\newtheorem{definition}{Definition}
\newtheorem{assumption}{Assumption}
\newtheorem{remark}{Remark}

\newcommandx{\rl}[2][1=]{\todo[linecolor=red,backgroundcolor=red!25,bordercolor=red,#1]{\textbf{Romain:} #2}}
\newcommandx{\rf}[2][1=]{\todo[linecolor=blue,backgroundcolor=blue!25,bordercolor=blue,#1]{\textbf{Raphael:} #2}}
\newcommandx{\ra}[2][1=]{\todo[linecolor=blue,backgroundcolor=blue!25,bordercolor=green,#1]{\textbf{Reda:} #2}}
\newcommandx{\td}[2][1=]{\todo[inline,size=\large,#1]{#2}}

\icmltitlerunning{Decentralized Exploration}

\begin{document}

\twocolumn[
\icmltitle{Decentralized Exploration in Multi-Armed Bandits - Extended version}



\icmlsetsymbol{equal}{*}

\begin{icmlauthorlist}
\icmlauthor{Rapha\"el F\'eraud}{o}
\icmlauthor{R\'eda Alami}{o}
\icmlauthor{Romain Laroche}{m}
\end{icmlauthorlist}

\icmlaffiliation{o}{Orange Labs}
\icmlaffiliation{m}{Microsoft Research}
\icmlcorrespondingauthor{Rapha\"el F\'eraud}{raphael.feraud@orange.com}



\vskip 0.3in
]



\printAffiliationsAndNotice{\icmlEqualContribution} 

\begin{abstract}
We consider the \textit {decentralized exploration problem}: a set of players collaborate to identify the best arm by asynchronously interacting with the same stochastic environment. 
The objective is to ensure privacy in the best arm identification problem between asynchronous, collaborative, and thrifty players. 
In the context of a digital service, we advocate that this decentralized approach allows a good balance between conflicting interests: the providers optimize their services, while protecting privacy of users and saving resources.
We define the privacy level with respect to the amount of information an adversary could infer by intercepting all the messages concerning a single user. We provide a generic algorithm {\sc Decentralized Elimination}, which uses any best arm identification algorithm as a subroutine. We prove that this algorithm ensures privacy, with a low communication cost, and that in comparison to the lower bound of the best arm identification problem, its sample complexity suffers from a penalty depending on the inverse square of the probability of the most frequent players. 
Then, thanks to the generality of the approach, we extend the proposed algorithm to the non-stationary bandits. Finally, experiments illustrate and complete the analysis.
\end {abstract}

\renewcommand{\thesection}{\arabic{section}}
\renewcommand{\thesubsection}{\arabic{section}.\arabic{subsection}}

\section {Introduction}

\subsection{Motivations}

\begin{figure}[ht]
  \centering
  {\includegraphics[width=9 cm]{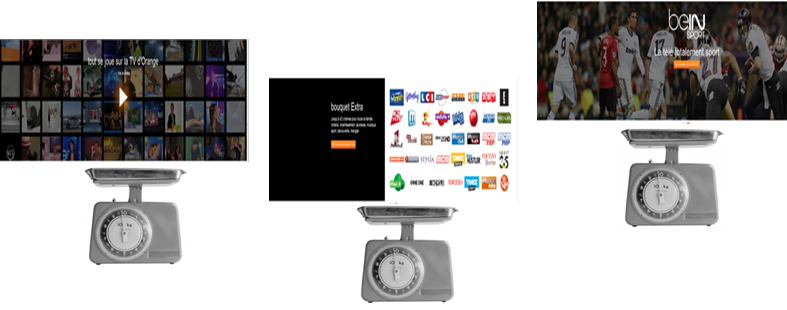}}
  \caption{ A/B testing - choosing the best message for promoting Orange TV. }
  \label{fig:marketing}
\end {figure}

For promoting their products in the digital world, most of companies perform marketing or advertising campaigns. For instance, when a cookie visits a web page, a banner containing marketing information is printed. If the cookie clicks on it, it is considered as a positive outcome. 
For maximizing the clicks on their banners, the companies need to carefully choose their messages. For instance for promoting Orange TV, should I highlight sport, series, movies,... (figure \ref {fig:marketing} ) ?  

For performing A/B testing, the interactions of the users with the service (application, web page...) and notably the clicks of the users are gathered by a server, and then are processed to find the best option. It works well, but this functional architecture (figure \ref {fig:architecture}) allow companies to log all interactions of all users: privacy cannot be guaranteed.

\begin{figure}[ht]
  \centering
  {\includegraphics[width=9 cm]{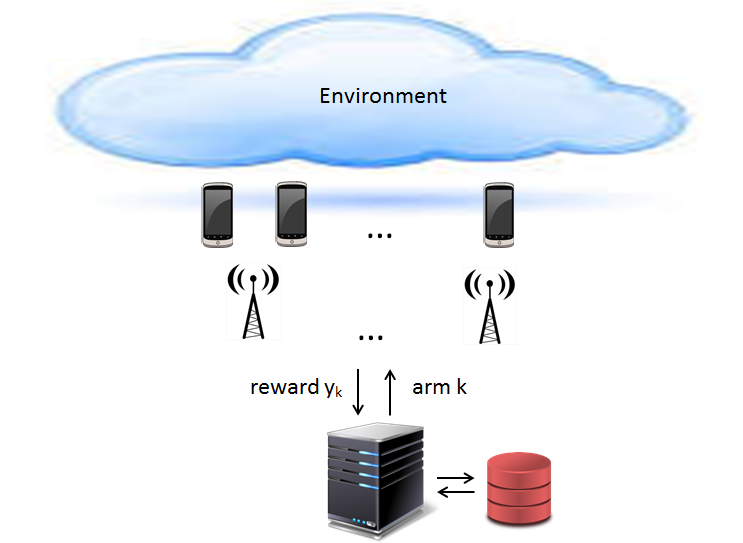}}
  \caption{A/B testing: functional architecture. }
  \label{fig:architecture}
\end {figure}

The proposed approach consists in performing decentralized exploration for finding the best arm with high probability. The application is run in the devices of the users and the interactions of the users stay in the devices, and hence privacy of users is ensured. The issue of this architecture is that the needed number of interactions for finding the best option is in the order of few hundreds. For a centralized server that shares the experiences of thousands of users it is small, but for a single user it is too much. That is why, we allow the devices to exchange messages (figure \ref {fig:architecture})

\begin{figure}[ht]
  \centering
  {\includegraphics[width=9 cm]{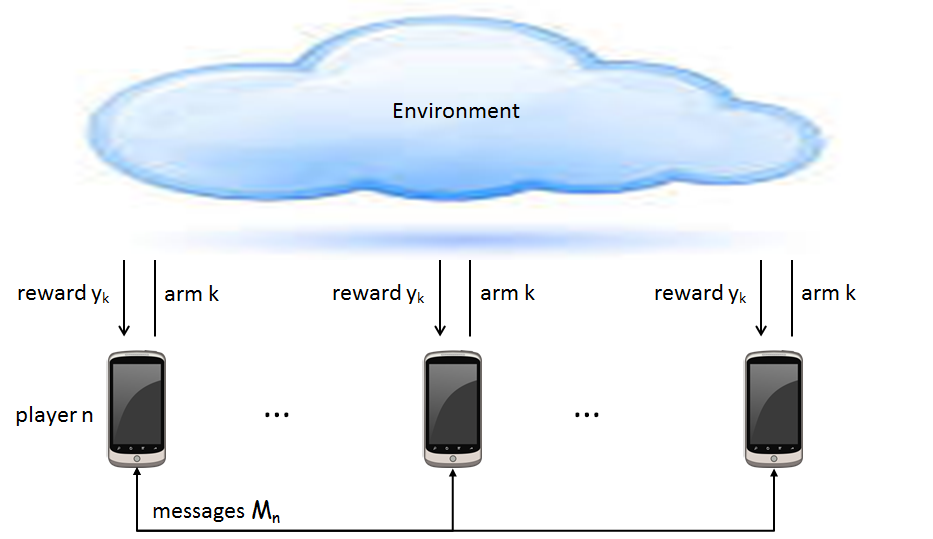}}
  \caption{Collaborative Exploration.}
  \label{fig:architecture}
\end {figure}

\subsection {Principle}

When the event {\it "player $n$ is active"} occurs, player $n$ reads the messages received from other players and then chooses an arm to play. 
The reward of the played arm is revealed to player $n$. Finally, she may send a message to the other players for sharing information about the arms.

The decentralized approach presents significant advantages. 
First, the clicks of users contain information that may be embarrassing when revealed, or that can be used by a third party in an undesirable way.

The decentralization of exploration favors privacy since the click stream is not transmitted. 
However, it is not sufficient. The messages sent by a user may still contain private information such as her favorite topics, and therefore her political views, sexual orientation...
As the players broadcast messages to other players, a malicious adversary can pretend to be a player, and then listening the exchanged messages.
To ensure privacy one must guarantee that no useful information can be inferred from the messages sent by a single user.
Second, the decentralization of exploration reduces the communication cost. 
This is a significant requirement for the Internet of Thing applications, since the smart devices often run on batteries.
Third and finally, for all digital applications and in particular for the mobile phone applications, the decentralization with a low communication cost increases the responsiveness of applications by minimizing the number of interactions between the application server and the devices.

Finally, the objective of the {\it decentralized exploration problem} is threefold: 
\begin{enumerate}
\vspace {-0.3 cm}
\item {\it sample efficiency}: finding a near-optimal arm with high-probability using a minimal number of interactions with the environment.
\vspace {-0.1 cm}
\item {\it user privacy}: protecting information contained in the interaction history of a single player.
\vspace {-0.1 cm}
\item {\it low communication cost}: minimizing the number of exchanged messages.
\vspace {-0.1 cm}
\end{enumerate}

\subsection {Related works}

The problem of the best arm identification has been studied in two distinct settings in the literature:
\begin{itemize}
\item the fixed budget setting: the duration of the exploration phase is fixed and is known by the forecaster, and the objective is to maximize the probability of returning the best arm \cite{BMS2009,AB2010,GGL2012};
\item the fixed confidence setting: the objective is to minimize the number of rounds needed to achieve a fixed confidence to return the best arm \cite{EMM2002,KTAS2012,GGL2012,KK2013}.
\end{itemize}

In this paper, we focus on the fixed confidence setting. Its theoretical analysis is based on the {\it Probably Approximately Correct} framework \cite{V1984}, and focuses on the sample complexity to identify a near-optimal arm with high probability. This theoretical framework has been used to analyze the best arm identification problem in \cite {EMM2002}, the dueling bandit problem in \cite {UCFN2013}, the batched bandit problem in \cite {PRCS2015}, the linear bandit problem in \cite {SLM2014}, the contextual bandit problem in \cite {FAUC2016}, and the non-stationary bandit problem in \cite {AFM2017}.

The {\it decentralized multi-player multi-armed bandits} have been studied for opportunistic spectrum access in \cite{LZ2010,AM2014,NKJ2018} or for optimizing communications in Internet of Things, even when no sensing information is available \cite{BK2018}. The objective is to avoid collisions between concurrent players that share the same channels, while choosing the best channels and minimizing the communication cost between players. 

Recent years have seen an increasing interest for the study of the distributed collaborative scheme, where there is no collision when players choose the same arm at the same time.
The distributed collaborative multi-armed bandits have been studied when the agents communicate through a neighborhood graph in \cite {SBHOJK2013,LSL2016}. 
Here, we allow each player to broadcast messages to all players. 
In \cite {CCDJ2017}, a team of agents collaborate to handle the same multi-armed bandit problem. At each step the agent can broadcast her last obtained reward for the chosen arm to the team or pull an arm. 
The communication cost corresponds to the lost of the potential reward. As the pull of the arm of the agent is broadcasted, this approach does not ensure privacy of users.
The tradeoff between the communication cost and the regret has been studied in the case of distributed collaborative non-stochastic experts \cite {KLR2012}. 
In \cite {HRPA2017}, the best arm identification task with fixed budget is distributed using Thompson Sampling in order to accelerate the exploration of the chemical space.
In \cite {HKKLS2013}, the best arm identification task with fixed confidence is distributed on a parallel processing architecture. 
The analysis focuses on the trade-off between the number of communication rounds and the number of pulls per player. 
Here, we consider here that the players activation is under the control of the environment. As a consequence, synchronized communication rounds can no longer be used to control the communication cost. In our paper, the cost of communications is assessed by the number of exchanged messages. 

Moreover, our purpose is also to protect privacy of players. In the current context of massive storage of personal data and massive usage of models inferred from personal data, privacy is an issue. Even if individual data are anonymized, the pattern of data associated with an individual is itself uniquely identifying.
The $k$-anonymity approach \cite {S2002} provides a guarantee to resist to direct linkage between stored data and the individuals. However, this approach can be vulnerable to composition attacks: an adversary could use side information that combined with the $k$-anonymized data allows to retrieve a unique identifier \cite {GKS2008}.  The {\it differential  privacy} \cite{DMNS2006}  provides an alternative approach. The sensitive data  are  hidden.  The  guarantee  is  provided  by  algorithms that allow to extract information from data. An algorithm is differentially private if the participation of
any  record  in  the  database  does  not  alter  the  probability  of any  outcome  by very much. The {\it differential  privacy} has been extended to {\it local differential  privacy} in which the data remains private even from the learner \cite {DJW2014}.
In \cite{GUK2018}, the authors propose an approach which handles the stochastic multi-armed bandit problem, while ensuring {\it local differential  privacy}. The $\epsilon$-differential privacy  is ensured to the players by using a stochastic corruption of rewards.  As all the rewards are transmitted to a centralized bandit algorithm, this approach has the maximum communication cost.
Here, we define the privacy level with respect to the information about the preferred arms of a player, that an adversary could infer by intercepting the messages of this player. The messages could be corrupted feedbacks as in \cite{GUK2018}, or as we choose a more compact representation of the same information.

\subsection {Our contribution}

In Section 2, we propose a new problem setting for ensuring privacy in the best arm identification problem between asynchronous, collaborative, and thrifty players. 
In Section 3, we propose a generic algorithm, {\sc Decentralized Elimination}, which handles the {\it decentralized exploration problem} using any best arm identification algorithm as a subroutine. Theorem 1 states that {\sc Decentralized Elimination} ensures privacy, finds an approximation of the best arm with high probability, and requires a low communication cost. Furthermore, Theorem 2 states a generic upper bound of the sample complexity of {\sc Decentralized Elimination}. More specifically, Corollary 1 and 2 state the sample complexity bound when respectively {\sc Median Elimination} and {\sc Successive Elimination} \cite{EMM2002} are used as subroutine. 
Then, in Section 4, we extend the algorithmic approach to the {\it decentralized exploration in non-stationary bandit problem}.
In Section 5, to illustrate and complete the analysis, we empirically compare the performances of {\sc Decentralized Elimination} with two natural baselines \cite {KLR2012}:
an algorithm that does not share any information between the players, and hence that ensures privacy with a zero communication cost, and a centralized algorithm that shares all the information between players, and hence that does not ensure privacy and that has the maximum communication cost. 

\section{The decentralized exploration problem}

Let $\mathcal {N}= \{1,...,N\}$ be a set of $N$ players. Let $x \in \mathcal {N}$ be a discrete random variable which realization denotes the index $n$ of the {\it active} player (the player for which an event occurs). Let $P_x$ be the probability distribution of $x$ which is  assumed to be stationary and unknown to the players. 
Let $\mathcal{K}=\{1,...,K\}$ be a set of $K$ arms.
Let $y^n_{k} \in [0,1]$ be the bounded random variable which realization denotes the reward of arm $k$ for player $n$, and $\mu^n_k$ be its mean reward. 
Let ${\bf y}_{x=n}=\{y^n_{k}\}_{k\in\mathcal{K}}$ be the vector of independent random variables $y^n_{k}$. Let $P_{{\bf y}}$ and $P_{x,{\bf y}}$ be respectively the probability distribution of ${\bf y}$ and the joint probability distribution of $x$ and ${\bf y}$, which are assumed to be unknown to the players. 

\begin {assumption}[stationary rewards]
 The mean reward of arms does not depend on time: $\forall t$, $\forall n$ $\in \mathcal {N}, $and $\forall k$ $\in \mathcal {K}$, $\mu^n_k(t)=\mu^n_k$.
 \label{assump:stationaryrewards}
\end {assumption}
\begin {assumption}[multi-armed bandits]
 The mean reward of arms does not depend on the player: $\forall n$ $\in \mathcal {N}$ and $\forall k$ $\in \mathcal {K}$, $\mu^n_k=\mu_k$.
 \label{assump:MAB}
\end {assumption}

Assumption \ref {assump:stationaryrewards} and \ref {assump:MAB} are used to focus on the stochastic multi-armed bandits. This section lays the theoretical foundations of the {\it decentralized exploration problem} in its elementary form. 
The next section proposes an extension to the {\it decentralized exploration in non-stationary bandits}. The extension to the {\it decentralized exploration in contextual bandits} is discussed in future works.

\begin {definition}[$\epsilon$-optimal arm]
An arm $k \in \mathcal{K}$ is said to be $\epsilon$-optimal, if $\mu_{k} \geq \mu_{k^*} - \epsilon$, where $k^*=\arg \max_{k \in \mathcal{K}} \mu_k$ and $\epsilon \in (0,1].$
$\mathcal {K}_\epsilon$ denotes the set of $\epsilon$-optimal arms.
\label {def:epsilonarm}
\end {definition}

\begin {definition}[message]
A message $\lambda_k^n \in \{0,1\}$ is a binary random variable, that is sent by player $n$ to other players, and where $\lambda_k^n=1$ means that player $n$ estimates that $k$ is not an $\epsilon$-optimal arm.\footnote {We choose a Bernoulli random variable for the sake of clarity. Notice that any random variable could be used as message.} 
\label {def:message}
\end {definition}

Let $\mathcal {M}_n$ be the set of sent messages by player $n$ at stopping time.
Let ${\mathcal {K}}^n(l^n) \subseteq  \mathcal{K}$ be the set of remaining arms at epoch $l^n \in \{1,...,L\}$ for player $n$, where $L$ is the maximal number of epochs.
 
\begin {definition}[$(\epsilon,\eta)$-private]
The decentralized algorithm $\mathcal {A}$ is $(\epsilon,\eta)$-private for finding an $\epsilon$-optimal arm, if for any player $n$, an adversary, that knows $\mathcal {M}_n$, the set of messages of player $n$, and the algorithm $\mathcal {A}$, cannot infer what arm is $\epsilon$-optimal for player $n$ with a probability higher than $1 -\eta$:
\begin {align*}
& \forall n \in \mathcal {N}, \forall l^n \in \{1,...,L\}, \nexists  \eta_1, 0 \leq \eta_1<\eta \leq 1, \\
& \mathds {P}\left(\mathcal{K}^n(l^n) \subseteq \mathcal {K}_\epsilon  | \mathcal {M}_n, \mathcal {A}\right) \geq 1- \eta_1.
\end {align*}
\label {def:private}
\end {definition}


$1-\eta$ is the confidence level associated to the decision of the adversary.
If $\eta$ is small, then the adversary can use the set of messages $\mathcal {M}_n$ to infer with high probability which arm is an $\epsilon$-optimal arm for player $n$.
If $\eta$ is high, the only information, that can be inferred by the adversary, is that the probability that an arm is an $\epsilon$-optimal of arm for player $n$ is a little bit higher than $0$, which can be much lesser than the random choice $1/K$. $\eta$ is a parameter which allows to tune the level of privacy: the higher $\eta$, the higher the privacy protection.

The goal of the {\it decentralized exploration problem} (see Algorithm \ref {DBA}) is to design an algorithm, that, when run on each player, samples effectively to find an $\epsilon$-optimal arm for each player, while ensuring $(\epsilon,\eta)$-privacy to players, and minimizing the number of exchanged messages.

\begin{algorithm}[!h]
  \caption{\sc Decentralized Exploration Problem}
  \label{DBA}
  {\bf Inputs:} $\mathcal {K}$, $\epsilon \in [0,1]$, $\eta \in [0,1]$\\
	{\bf Output:} an arm in each set $\mathcal{K}^n(l^n)$\\
    {\bf Initialization:} $l^n:=1$, $\mathcal{K}^n(l^n):=\mathcal{K}$\\
    \vspace {-0.3 cm}
\begin{algorithmic} [1]
		\REPEAT
            \STATE a player is sampled: $n \sim P_x$
			\STATE player $n$ gets the messages of other players 
			\STATE arm $k \in \mathcal{K}^n(l^n)$ is played by player $n$
			\STATE player $n$ receives reward $y^n_k \sim P_{x=n,y}$
            \STATE {\bf if} player $n$ updates $\mathcal{K}^n(l^n)$ {\bf then} $l^n:=l^n+1$
      		\STATE player $n$ sends a message to other players
		\UNTIL {$\left( \forall n  \in \mathcal {N}, | \mathcal{K}^n(l^n)|=1\right)$ }
\end{algorithmic}
\end{algorithm}

The lower bound of the number of samples in $P_{x,\bf y}$ needed to find with high probability an $\epsilon$-optimal arm, which is $\Omega\left( \frac {K}{\epsilon^2} \log \frac {1}{\delta}\right)$ \cite{MT2004}, holds for the {\it decentralized exploration problem}, since a message can be sent at each time an arm is sampled by a player. 
The number of messages, that has to be exchanged in order to find with high probability an $\epsilon$-optimal arm, could be zero if each player independently handles the best arm identification problem.

\begin {assumption}[all players are active]
 $\forall n$ $\in \mathcal {N}$, $P_x(x=n) \neq 0$.
 \label{assump:active}
\end {assumption}

Assumption \ref {assump:active} is a sanity check assumption for the {\it decentralized exploration problem}. Indeed, if it exists a player $n$ such that $P_x(x=n) = 0$, then Algorithm \ref {DBA} never stops (the stopping condition line 8 never happens).

\section {Decentralized Elimination}

\subsection {ArmSelection subroutine}

Before describing a generic algorithm for the {\it decentralized exploration problem}, we need to define an $\operatorname{ArmSelection}$ subroutine that handles all best arm identification algorithms.
Let $\overline {\mathcal{K}^n}(l^n)$ and ${\mathcal {K}^n}(l^n)$ be respectively the set of eliminated arms and the set of remaining arms of player $n$ at elimination epoch $l^n$, such that $\overline {\mathcal{K}^n}(l^n) \cup {\mathcal {K}^n}(l^n) = {\mathcal {K}^n}(l^n-1)$.

\begin {definition}[ArmSelection subroutine]
 an $\operatorname{ArmSelection}$ subroutine takes as parameters an approximation factor $\epsilon$, a confidence level $1-\eta$, and a set of remaining arm ${\mathcal {K}^n}(l^n)$. It samples a remaining arm in ${\mathcal {K}^n}(l^n)$ and returns the set of eliminated arms  $\overline {\mathcal{K}^n}(l^n)$. An $\operatorname{ArmSelection}$ subroutine satisfies Properties 1 and 2.
 \label {def:subroutine}
 \end {definition}

Let $t^n$ be the number of calls of the $\operatorname{ArmSelection}$ subroutine. Let $\mathcal{H}_{t^n}$ be the sequence of rewards of chosen arms $\{(k_1,y^n_{k_1}),(k_2,y^n_{k_2}),...,(k_{t^n},y^n_{k_{t^n}})\}$. Let $f: \{1,...,L\} \rightarrow [0,1]$ be a function such that $\sum_{l^n=1}^{L} f(l^n)=1$. 

\paragraph {Property 1. (remaining $\epsilon$-optimal arm)}

\begin{align*}
& \forall l^n \in \{1,...,L\}, \mathcal{K}^n \left(l^n \right)  \subset \mathcal{K}^n\left(l^n-1 \right),\\
& \mathds{P} \left( \{ \mathcal {K}^n(l^n) \cap \mathcal {K}_\epsilon=\emptyset \} | \mathcal{H}_{t^n-1} , \mathcal{K}^n \left(l^n -1 \right)  \cap \mathcal{K}_\epsilon \neq \emptyset \right) \leq \eta \times f(l^n).
\end{align*}

\paragraph {Property 2. (finite sample complexity)}

\begin{align*}
& \forall \eta \in (0,1), \forall \epsilon \in (0,1], \exists t^n \geq 1,\\
& \mathds{P} \left( \{ \mathcal {K}^n(L) \subset {\mathcal {K}_\epsilon} \}  | \mathcal{H}_{t^n-1}  \right) \geq 1- \eta.
\end{align*}

Property 1 ensures that with high probability at least an $\epsilon$-optimal arm remains in the set of arms $\mathcal {K}^n(l^n)$, while Property 2 ensures that the $\operatorname{ArmSelection}$ subroutine finds in a finite time an $\epsilon$-optimal arm whatever the confidence level $1-\eta$ and the approximation factor $\epsilon$. 
To the best of our knowledge, all best arm identification algorithms
can be used as  $\operatorname{ArmSelection}$ subroutine with straightforward transformations. We consider three classes of best arm identification algorithms.

\paragraph {The fixed-design algorithms} use {\sl uniform sampling} during a predetermined number of samples. {\sc Naive Elimination} ($L=1$ and $f(l^n)=1$) and {\sc Median Elimination} ($L=\log_2 K$ and $f(l^n)=1/2^{l^n}$) \cite {EMM2002} are {\it fixed-design} algorithms which can be used as $\operatorname{ArmSelection}$ subroutines.

\paragraph {The successive elimination algorithms} are based on {\sl uniform sampling} and {\sl arm eliminations}. At each time step a remaining arm is uniformly sampled. The empirical mean of the played arm is updated. The arms, which cannot be an $\epsilon$-optimal arm with high probability, are discarded. If suboptimal arms are discarded the epoch $l$ is increased by one.
{\sc Successive Elimination} ($L=K$ and $f(l^n)=1/K$) \cite {EMM2002}, {\sc KL-Racing} ($L=K$ and $f(l^n)=1/K$) \cite {KK2013} are {\it successive elimination} algorithms which can be used as $\operatorname{ArmSelection}$ subroutines.

\paragraph {The explore-then-commit algorithms} are based on {\it adaptive sampling} and {\it a stopping rule}. Rather than choosing arms uniformly, the {\it explore-then-commit} algorithms play one of the two critical arms: the empirical best arm, and the empirical suboptimal arm associated with the maximum upper confidence bound. The stopping rule simply tests if the difference, between the maximum  of upper confidence bound of suboptimal arms and the lower confidence bound of the empirical best arm, is higher than the approximation factor $\epsilon$. When the algorithm stops it returns the best arm.
{\sc LUCB} \cite {KTAS2012}, {\sc KL-LUCB} \cite {KK2013}, {\sc UGapEc} \cite {GGL2012} can also be used as $\operatorname{ArmSelection}$ subroutines by returning the set of eliminated arms when the stopping event occurs ($L=1$ and $f(l^n)=1$).

\subsection{Algorithm description}

The basic idea of {\sc Decentralized Elimination} is to use the vote of independent players, 
which communicate the arm they would like to eliminate with a high probability of failure for ensuring privacy. As the players are independent, the probability of failure of the vote is the multiplication of the individual probability of failures. The number of players needed for eliminating an arm is provided by the analysis.

{\sc Decentralized Elimination} (see Algorithm \ref {DSE}) takes as parameters the privacy level $\eta$, the failure probability $\delta$, the approximation factor $\epsilon$, and an $\operatorname{ArmSelection}$ subroutine. It outputs an $\epsilon$-optimal arm for each player with high probability. The algorithm sketch is described below.
\paragraph {}
When player $n$ is active (i.e. when player $n$ is sampled):
\begin {itemize}
\item player $n$ gets messages from other players  (line 3).
\item When enough players have eliminated an arm, it is eliminated from the shared set of arms $\mathcal {K}(l)$ and from the set of arms $\mathcal {K}^n(l^n)$ of player $n$ with a low probability of failure (lines 5-10). 
\item When there is only one arm in $\mathcal {K}(l)$, it is an $\epsilon$-optimal arm with high probability $1-\delta$, and the set of arms of player $n$ is $\mathcal {K}(l)$ (line 11).

\item An $\operatorname{ArmSelection}$ subroutine, run with a low confidence level $1-\eta$ (i.e. high privacy level) on the set $\mathcal {K}^n(l^n)$, samples an arm and returns $\overline {\mathcal{K}}^n(l^n)$ the set of arms that player $n$ has eliminated at step $t^n$ (line 13).

\item When player $n$ has eliminated an arm, she communicates to other players the index of the arm (lines 14-20). 
\end {itemize}

\begin{algorithm}[!h]
\caption {\sc Decentralized Elimination}
\label {DSE}
{\bf Inputs:} $\epsilon \in (0,1]$, $\eta \in (0,1)$, $\delta \in [\eta^N,\eta^2]$, $\mathcal {K}$, an $\operatorname{ArmSelection}$ subroutine\\
{\bf Output:} an arm in each set $\mathcal{K}^n(l^n)$\\
{\bf Initialization:}  $l:=1$, $\mathcal{K}(l):=\mathcal{K}$, $\forall n$ $t^n:=1$, $l^n:=1$,  $\mathcal{K}^n(l^n):=\mathcal{K}$, $ \forall \left(k,n \right) \  \lambda^n_k:=0$\\ 
\vspace{-0.5 cm}
\begin{algorithmic} [1]
\REPEAT
	\STATE player $n$ is sampled: $n \sim P_x$
	\STATE player $n$ gets the messages $\lambda_k^j$ from other players
     \IF { $|\mathcal {K}(l)|>1$ } 
    \FOR {all $k \in \mathcal{K}(l)$}
        \IF { $\sum_{j=1}^N \lambda^j_k \geq  {\lfloor \frac {\log \delta}{\log \eta} 	\rfloor}$} 
        \STATE $\mathcal {K}(l):=\mathcal {K}(l) \setminus \{k\}$, $l:=l+1$
        \STATE $\mathcal {K}^n(l^n):=\mathcal {K}^n(l^n) \setminus \{k\}$
        \ENDIF
         \ENDFOR
        \STATE {\bf else} $\mathcal {K}^n(l^n):=\mathcal {K}(l)$
        \ENDIF
	\STATE $\overline {\mathcal{K}}^n(l^n):=\operatorname{ArmSelection}(\epsilon,\eta,\mathcal{K}^n(l^n))$  
    \IF {$|\overline {\mathcal{K}}^n(l^n)| > 1$}
    \STATE $l^n:=l^n+1$
    \FOR {all $k \in \overline {\mathcal{K}}^n(l^n)$}
			\STATE $\mathcal {K}^n(l^n):=\mathcal {K}^n(l^n) \setminus \{k\}$ 
            \STATE $\lambda_k^n:=1$, $\lambda_k^n$ is sent to other players
	\ENDFOR
    \ENDIF
    \STATE $t^n:=t^n+1$
\UNTIL {$\forall n$  $|\mathcal {K}^n (l^n)| = 1$}
\end{algorithmic}
\end{algorithm} 

\subsection {Analysis of the algorithm}

Theorem~1 states the upper bound of the communication cost for obtaining with high probability an $\epsilon$-optimal arm while ensuring $(\epsilon,\eta)$-privacy to the players. The communication cost depends only on the problem parameters: the privacy constraint $\eta$, the probability of failure $\delta$, the number of actions, and notably not on the number of samples.  Notice that the probability of failure is low since the failure probability is lower than the level of privacy guarantee: $\delta < \eta$. 

\begin {theorem}[]
{\it Using any $\operatorname{ArmSelection}$ subroutine,  {\sc Decentralized Elimination} is an $(\epsilon,\eta)$-private algorithm, that finds an $\epsilon$-optimal arm with a failure probability $\delta \leq \eta^{\lfloor \frac {\log \delta}{\log \eta} \rfloor}$ and that exchanges at most $\lfloor \frac {\log \delta}{\log \eta} \rfloor K -1$ messages.}
\label{theorem:private}
\end {theorem}

To finely analyze the sample complexity of {\sc Decentralized Elimination} algorithm, one needs to handle the randomness of the voting process. Let $T_{P_{x,\bf y}}$ be the number of samples in $P_{x,\bf y}$ when {\sc Decentralized Elimination} stops with high probability.
Let $T_{P_{\bf y}}$ be the number of samples in $P_{\bf y}$ needed by the 
$\operatorname{ArmSelection}$ subroutine to find an 
$\epsilon$-optimal arm with high probability.
Let $\mathcal {N}_M$ be the set of the $M=\lfloor \frac {\log \delta}{\log \eta} \rfloor$ most likely players, let $p^* = \min_{n \in \mathcal {N}_M} P_x (x=n)$, and let $p^\dagger = \min_{n \in \mathcal {N}} P_x (x=n)$.

\begin {theorem}[]
{\it Using any $\operatorname{ArmSelection}$ subroutine, with a probability higher than \\ $(1-\delta) \left(1-I_{1-p^*}\left(T_{P_{x,\bf y}}-T_{P_{\bf y}}, 1+ T_{P_{\bf y}}\right)\right)^{\lfloor \frac {\log \delta}{\log \eta} \rfloor}$ {\sc Decentralized Elimination} stops after at most:}
\begin {displaymath}
T_{P_{x,{\bf y}}} =
\frac{T_{P_{\bf y}}}{p^*}
+\frac {1}{(p^*)^2} \sqrt{\frac{p^*T_{P_{\bf y}}}{2}\log \frac {1}{\delta}} +
\frac{1}{2(p^*)^2} \log \frac {1}{\delta}
\end {displaymath}
{\it  samples in $P_{x,\bf y}$, where $I_a(b,c)$ denotes the incomplete beta function evaluated at $a$ with parameters $b,c$.}
\label {theorem:sample}
\end {theorem}

As the number of players involved in the vote is set as small as possible $\lfloor \frac {\log \delta}{\log \eta} \rfloor$, Theorem 2 provides with high probability \footnote{for instance, $I_{0.99}(500,500)=1.47\times10^{-302}$} the sample complexity of {\sc Decentralized Elimination}. Notice, that when the number of players is high, and when the distribution of players is far from the uniform distribution, we have $p^* \gg p^\dagger$. 

\begin {corollary}[]
{\it With a probability higher than $(1-\delta) \left(1-I_{1-p^*}\left(T_{P_{x,\bf y}}-T_{P_{\bf y}}, 1+ T_{P_{\bf y}}\right)\right)^{\lfloor \frac {\log \delta}{\log \eta} \rfloor}$ {\sc Decentralized Median Elimination}  stops after:}
\begin {displaymath}
\mathcal {O}\left(\left(   
\frac {K}{\lfloor \frac {\log \delta}{\log \eta} \rfloor \epsilon^2} 
+
\frac{1}{p^*}\right) \frac{1}{p^*} \log \frac {1}{\delta} \right)
 \text {\it samples in } P_{x,\bf y}.
\end {displaymath}
\label {coro:median}
\end {corollary}

\begin {corollary}[]
 {\it With a probability higher than $(1-\delta) \left(1-I_{1-p^*}\left(T_{P_{x,\bf y}}-T_{P_{\bf y}}, 1+ T_{P_{\bf y}}\right)\right)^{\lfloor \frac {\log \delta}{\log \eta} \rfloor}$  {\sc Decentralized Successive Elimination}  stop after:}
\begin {align*}
\mathcal {O} \left( 
\left(   
\frac {K}{\lfloor \frac {\log \delta}{\log \eta} \rfloor \epsilon^2} 
+
\frac{1}{p^*}\right)\frac {1}{p^*}\log \frac {1}{\delta}
+\frac {K}{p^*\epsilon^2} \log K \sqrt {\log \frac {1}{\delta}}
\right)
\end {align*}
{\it samples in } $P_{x,\bf y}$.
\label {coro:successive}
\end {corollary}

Corollaries \ref {coro:median} and \ref {coro:successive} state the number of samples in $P_{x,\bf y}$ needed to find an $\epsilon$-optimal arm by {\sc Decentralized Elimination} using respectively {\sc Median Elimination} and {\sc Successive Elimination} as $\operatorname {ArmSelection}$ subroutines. 

\begin {remark}
In the case of the uniform distribution of players, with a failure probability at most $\delta=\eta^N$ the number of sample in $P_{x,\bf y}$ needed by {\sc Decentralized Median Elimination} to find an $\epsilon$-optimal arm is:
\begin {displaymath}
\mathcal {O}\left( \frac {K}{\epsilon^2} \log \frac {1}{\delta} +N^2 \log \frac {1}{\delta}\right) \text { samples in } P_{x,\bf y}.
\end {displaymath}

In comparison to an optimal best arm identification algorithm, which communicates all the messages and does not provide privacy protection guarantee, which has a sample complexity in $\mathcal {O}\left(\frac {K}{\epsilon^2}\log \frac{1}{\delta} \right)$, the sample complexity of {\sc Decentralized Elimination} mostly suffers from a penalty depending on the inverse of the probability of the most frequent players, that in the case of uniform distribution of players is quadratic with respect to the number of players.
\end {remark}

\section {Decentralized exploration in non-stationary bandits}

Recently, the best arm identification problem has been studied in the case of non-stationary bandits, where Assumption \ref {assump:stationaryrewards} does not hold \cite {AFM2017,ABGMV2018}. 
In the first reference, the authors analyze the non-stationary stochastic best-arm identification in the fixed confidence setting by splitting the game into independent sub-games where the best arm does not change. 
In the second reference, the authors propose a simple and anytime algorithm, which is analyzed for stochastic and adversarial rewards in the case of fixed budget setting. 
For the consistency of the paper, which focuses on fixed confidence setting, we choose to extend {\sc Decentralized Elimination} to {\sc Successive Elimination} with {\sc Randomized Round-Robin} ({\sc SER3} \cite{AFM2017}). Basically, {\sc SER3} consists in shuffling the set of arms at each step of {\sc Successive Elimination}. {\sc SER3} works for the sequences where Assumption \ref {assump:positive} holds.

\begin {assumption}[Positive mean gap]
\label {assump:positive}
For any $k \in \mathcal {K} \setminus \{k^*\}$ and any $[\tau] \in \mathbb{T}(\tau)$ with $\tau \geq  \log \frac {K}{\eta}$, we have:
\begin{align*}
\Delta^*_k\left( [\tau]\right) =\frac{1}{\tau} \sum^\tau_{i=1} \sum_{j = i }^{i + K_i- 1} \frac{\Delta_{k^*,k}(j)}{K_i} > 0 , 
\end{align*}
where $\mathbb{T}(\tau)$ is the set containing all possible realizations of $\tau$ round-robin steps, $\Delta_{k^*,k}(t)$ is the difference between the mean reward of the best arm and the mean reward of arm $k$ at time $t$, and $K_t$ is the number of remaining arms at time $t$.
\end {assumption}

We provide below the sample complexity bound of {\sc Decentralized Successive Elimination} with {\sc Randomized Round-Robin} ({\sc DSER3}), which is simply {\sc Decentralized Elimination} using {\sc SER3} as the ArmSelection subroutine.

\begin {theorem}[]
\label {th:DSER3}
{\it For  $K\geq2$,  $\delta \in (0,0.5]$, $\epsilon \geq \delta/K$ for the sequences of rewards where Assumption \ref {assump:positive} holds, DSER3 is an $(\epsilon,\eta)$-private algorithm, that exchanges at most $\lfloor \frac {\log \delta}{\log \eta} \rfloor K -1$ messages, that finds an $\epsilon$-optimal arm with a probability at least $(1-\delta) \left(1-I_{1-p^*}\left(T_{P_{x,\bf y}}-T_{P_{\bf y}}, 1+ T_{P_{\bf y}}\right)\right)^{\lfloor \frac {\log \delta}{\log \eta} \rfloor}$, and that stops after: }
\begin {align*}
\mathcal {O} \left( 
\left(   
\frac {K}{\lfloor \frac {\log \delta}{\log \eta} \rfloor \epsilon^2} 
+
\frac{1}{p^*}\right)\frac {1}{p^*}\log \frac {1}{\delta}
+\frac {K}{p^*\epsilon^2} \log K \sqrt {\log \frac {1}{\delta}}
\right)
\end {align*}
{\it samples in } $P_{x,\bf y}$.
\end {theorem}

\begin{figure}[ht]
  \centering
  {\includegraphics[width=7 cm]{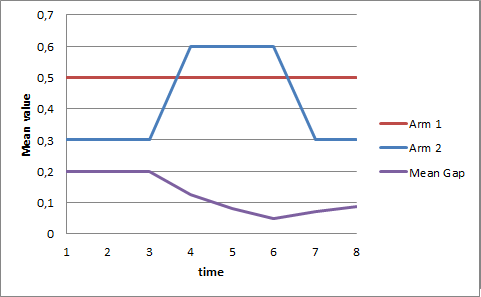}}
  \caption{Mean gap versus time. Assumption \ref {assump:positive} holds: the mean gap stays positive. }
  \label{Gap1}
\end {figure}

\begin{figure}[ht]
  \centering
  {\includegraphics[width=7 cm]{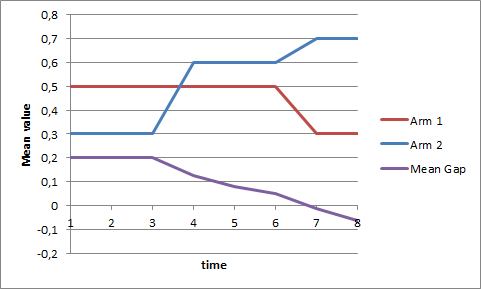}}
  \caption{Mean gap versus time. Assumption \ref {assump:positive} does not hold: a switch occurs a time $7$.}
  \label{Gap2}
\end{figure}

Finally {\sc Decentralized Successive Elimination} with {\sc Randomized Round-Robin and Reset} ({\sc DSER4}) handles any sequence of rewards: when Assumption \ref {assump:positive} does not hold a switch occurs (see figures \ref {Gap1}, \ref {Gap2}). 

{\sc DSER4} consists in using {\sc SER4} \cite{AFM2017} as the ArmElimination subroutine in {\sc Decentralized Elimination}. In addition, when a reset occurs in {\sc SER4}, {\sc Decentralized Elimination} is reset. 

\begin {theorem}[]
\label {th:DSER4}
{\it For  $K\geq2$,  $\epsilon \geq \frac {\eta}{K}$, $\varphi \in (0,1]$, $\epsilon \geq \delta/K$ for any sequences of rewards, {\sc DSER4} is an $(\epsilon,\eta)$-private algorithm, that exchanges on average at most $\varphi T(\lfloor \frac {\log \delta}{\log \eta} \rfloor K -1)$ messages, and that plays, with an expected probability at most $\delta+\varphi T I_{1-p^*}\left(T_{P_{x,\bf y}}-T_{P_{\bf y}}, 1+ T_{P_{\bf y}}\right)^{\lfloor \frac {\log \delta}{\log \eta} \rfloor}$, a suboptimal arm on average no more than: }
\begin {align*}
 \mathcal {O} \left( \frac{1}{\epsilon^2} \sqrt{\frac{\left(   
\frac {K}{p^*} \log K +
\frac {K}{\lfloor \frac {\log \delta}{\log \eta} \rfloor } 
\right)\frac {S}{p^*}\log \frac {1}{\delta}}{\delta} }
 +\frac {1}{(p^*)^2}\log \frac {1}{\delta}
\right),
\end {align*}
{\it times, where $S$ is the number of switches of best arms, $\varphi$ is the probability of reset in {\sc SER4}, $T$ is the time horizon, and the expected values are taken with respect to the randomization of resets.}
\end {theorem}

\section {Experiments}

\begin{figure*}[ht]
  \centering
  \subfigure[Problem 1 - Sample Complexity]{\includegraphics[width=7 cm]{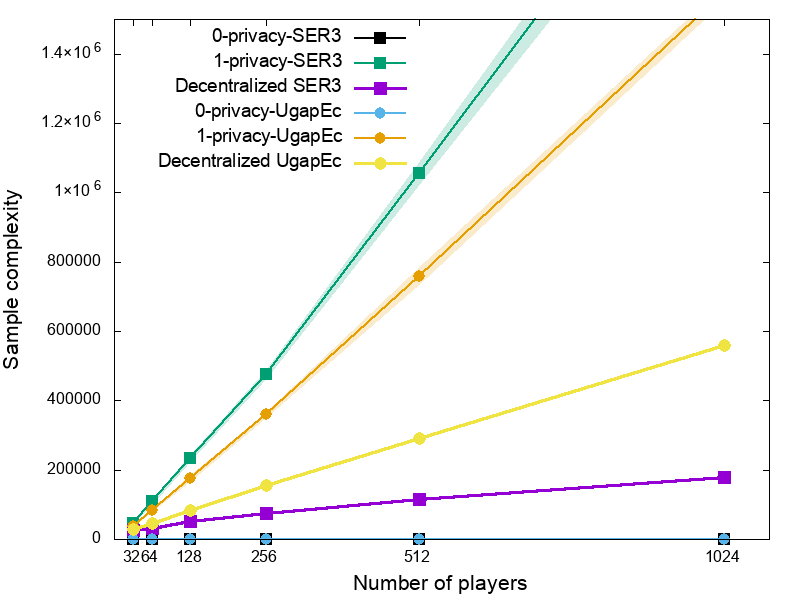}}
  \hspace{1pt}
  \subfigure[Problem 1 - Number of messages]{\includegraphics[width=7 cm]{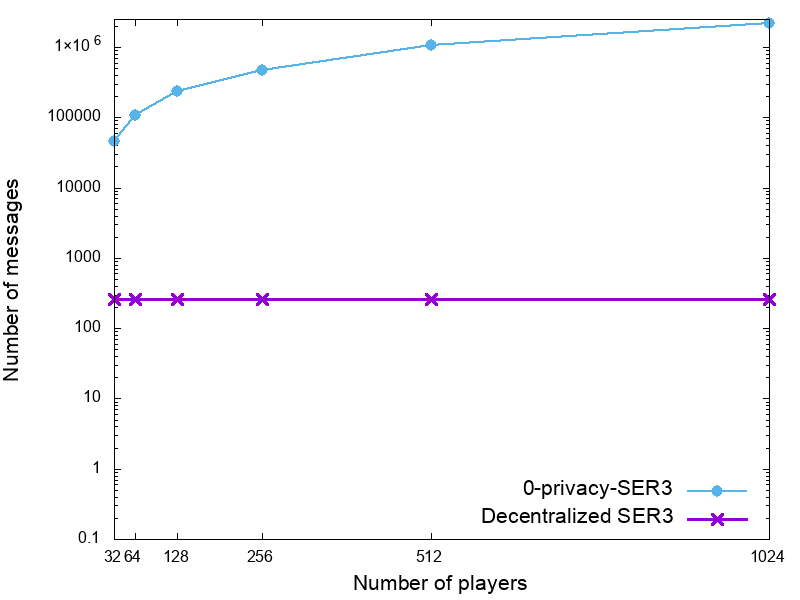}}
  \caption{Uniform distribution of players. The sample complexities of {\sc $0$-privacy} baselines are the same: $800$.}
  \label{Toy1}
\end{figure*}

\begin{figure*}[ht]
  \centering
  \subfigure[Problem 2 - Sample Complexity]{\includegraphics[width=7 cm]{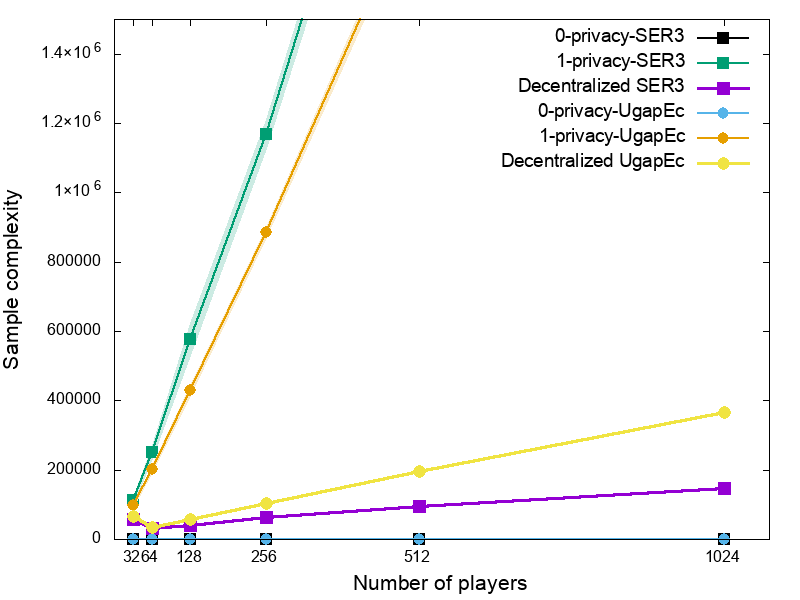}}
  \hspace{1pt}
  \subfigure[Problem 3 - Sample Complexity]{\includegraphics[width=7 cm]{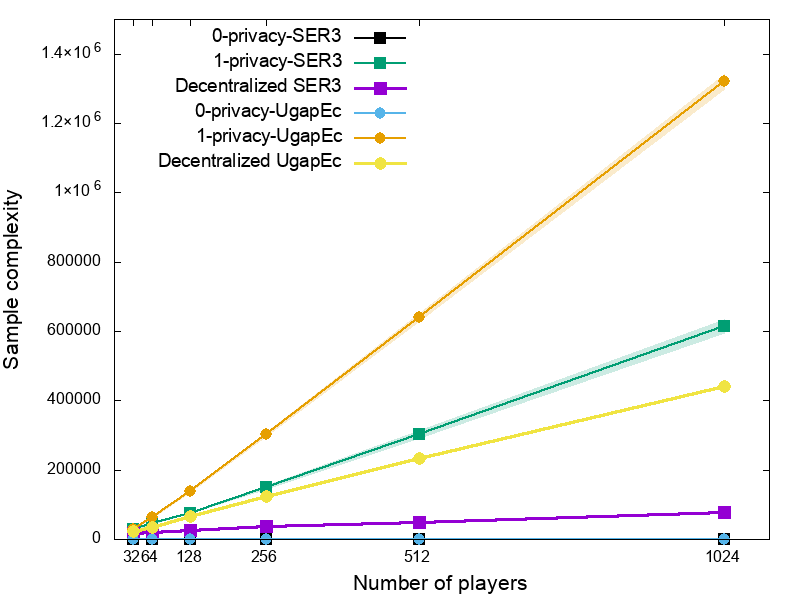}}
  \caption {$50 \%$ of players generates $80 \%$ of events (a), and the mean rewards of suboptimal arms linearly decrease (b).}
  \label{Toy2}
\end{figure*}

\subsection {Experimental setting}

To illustrate and complete the analysis of {\sc Decentralized Elimination}, we run three synthetic experiments:

\begin{itemize}
\item \textbf{Problem 1: Uniform distribution of players.} There are $10$ arms. The optimal arm has a mean reward $\mu_1=0.7$, the second one $\mu_2=0.5$, the third one $\mu_3=0.3$, and the others have a mean reward of $0.1$. Each player has a probability equal to $1/N$. 
\item \textbf{Problem 2: $50 \%$ of players generates $80 \%$ of events.}
The same $10$ arms are reused with an unbalanced distribution of players. The players are split in two groups of sizes $N/2$. When a player is sampled, a uniform random variable $z \in [0,1]$ is drawn. If $z < 0.8$  the player is uniformly sampled from the first group, otherwise it is uniformly sampled from the second group.
\item \textbf{Problem 3: non-stationary rewards.}
The distribution of players is uniform.  The same $10$ arms are reused.
The mean reward of the optimal arm does not change during time. The mean reward of suboptimal arms linearly decrease: $\mu(t)=\mu(0)-10^{-5}t$.
\end{itemize}

As comparison points, we include two natural baselines:
\begin {itemize}
\item \textsc{$1$-privacy}: an $(\epsilon,1)$-private algorithm that does not share any information between the players, and hence that runs at a zero communication cost. The ArmSelection subroutine is run with parameters $(\epsilon,\delta/N)$ to ensure that all the players find with a probability $1-\delta$ an $\epsilon$-optimal arm.
\item \textsc{$0$-privacy}: an $(\epsilon,0)$-private algorithm that shares all the information between players, and hence that runs at a minimal privacy and a maximal communication cost. 
This algorithm does not meet the original goal but is interesting as a reference to assess the sample efficiency loss stemming from the privacy constraint.
\end {itemize}
As {\sc ArmSelection} subroutines, We choose two frequentist algorithms
based on Hoeffding inequality: a {\it explore-then-commit} algorithm {\sc UGapEc} \cite {GGL2012} and a {\it successive elimination} algorithm {\sc SER3} \cite {AFM2017}, which handles non-stationary rewards. 
Combining {\sc Decentralized Elimination} and the two baselines with the two {\sc ArmSelection} subroutines, we compare $6$ algorithms ({\sc Decentralized SER3}, {\sc Decentralized UGapEc}, \textsc{$1$-privacy-SER3}, \textsc{$1$-privacy-UGapEc}, \textsc{$0$-privacy-SER3}, \textsc{$0$-privacy-UGapEc}) on the three problems.
The algorithms are compared with respect to two key performance indicators: the sample complexity and the communication cost. 
For all the experiments, $\epsilon$ is set to $0.25$, and $\delta$ is set to $0.05$. 
The privacy level $\eta$ is set to $0.9$. 
All the curves and the measures are averaged over $20$ trials.

\subsection {Results}

The results reveal that the sample efficiency of \textsc{$1$-privacy} baselines is horrendous on both problems: it increases super-linearly  
as the number of players increases. Worse, when the distribution of players moves away from the uniformity, which is the case in most of digital applications, the performances of \textsc{$1$-privacy} baselines decreases (see Figure \ref {Toy1}a, \ref {Toy2}a).
Contrary to \textsc{$1$-privacy} baselines, the performances of {\sc Decentralized UGapEc} and  {\sc Decentralized SER3} increases in Problem 2 (see Figure \ref {Toy2}a).
More precisely, the sample complexity curves of {\sc Decentralized UGapEc} and {\sc Decentralized SER3} exhibit two regimes: first the sample complexity decreases (between $32$ to $64$ players), and then the sample complexity linearly increases with the number of players. 
The values of hyper-parameters: $\delta=0.05$ and $\eta=0.9$, imply that the number $M={\lfloor \frac {\log \delta}{\log \eta} 	\rfloor}$ of player votes required to eliminate an arm is $28$. 
In Problem $2$ with $32$ players, it means that 
the algorithm has to wait for infrequent players votes to terminate. When the number of players is $64$, this issue disappears. This is the reason why the sample complexity for $64$ players is lower than for $32$ players. 

Concerning the {\sc ArmSelection} subroutines, we observe that \textsc{$1$-privacy-UGapEc} clearly outperforms \textsc{$1$-privacy-SER3} on stationary problems (see Figures \ref {Toy1}a and \ref {Toy2}a). Moreover, the performance gain of \textsc{$1$-privacy-UGapEc} increases with the number of players. This is due to the adaptive sampling strategy of {\sc UGapEc}: by sampling alternatively the empirical best arm and the most loosely estimated suboptimal arm, \textsc{$1$-privacy-UGapEc} reduces the variance of the sample complexity, and thus reduces the maximum of sample complexities of players. However, when used as a subroutine in {\sc Decentralized Elimination}, the {\it successive elimination} algorithms such as {\sc SER3} are more efficient: thanks to the different suboptimal arms which are progressively eliminated by different groups of voting players, \textsc{Decentralized SER3} clearly outperforms \textsc{Decentralized UGapEc} (see Figure \ref {Toy1}a and \ref {Toy2}a). 

When the mean rewards of suboptimal arms are decreasing (Figure \ref {Toy2}b), in comparison to {\sc SER3} the performances of {\sc UGapEc}, which is not designed for non-stationary rewards, collapse:
\textsc{$1$-privacy-UGapEc} and \textsc{Decentralized UGapEc} are respectively outperformed by \textsc{$1$-privacy-SER3} and \textsc{Decentralized SER3}. The optimistic approach used in the sampling rule of \textsc{UGapEc} is too optimistic when the mean reward are decreasing.

The communication cost is the number of exchanged messages: \textsc{$1$-privacy} baselines send zero messages,  while \textsc{$0$-privacy} baselines send $N-1$ messages per time step 
until the $\epsilon$-optimal arm is found. {\sc Decentralized SER3} needs three to four orders of magnitude less messages than \textsc{$0$-privacy-SER3} (see Figure \ref {Toy1}b).

\begin{figure*}[ht]
  \centering
  \subfigure[Problem 1:Uniform distribution of players]{\includegraphics[width=7 cm]{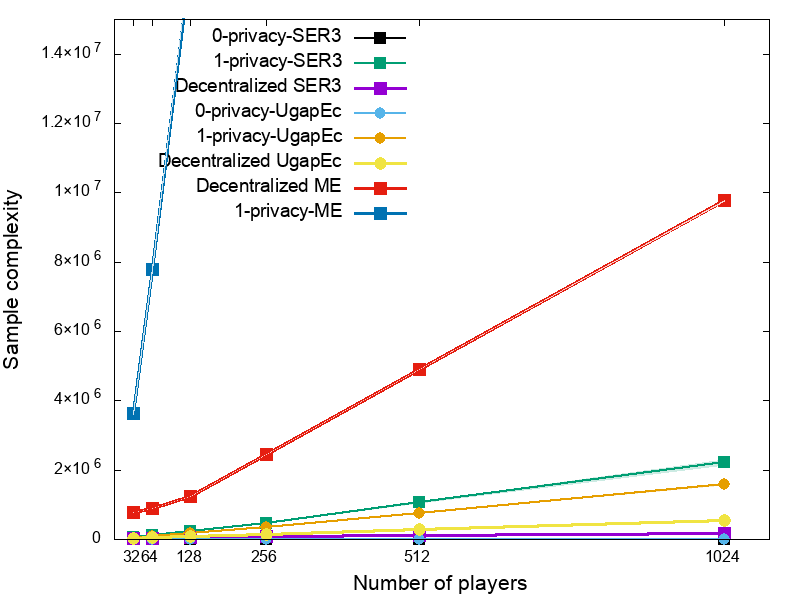}}
  \hspace{1pt}
  \subfigure[Problem 2: $50 \%$ of players generates $80 \%$ of events]{\includegraphics[width=7 cm]{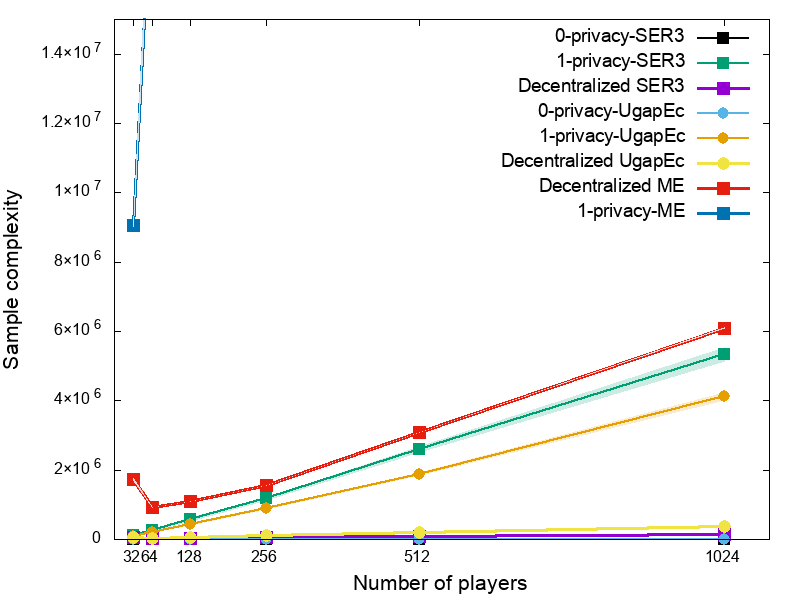}}
  \caption {Decentralized Median Elimination}
  \label{ME2}
\end{figure*}

Finally, {\sc Median Elimination} is designed to be order optimal in the worst case: its sample complexity is in $O(K \log \frac {1}{\delta})$. 
However, in practice it is clearly outperformed by {\sc Successive Elimination} or {\sc UGapEc} on both problems (see Figures \ref {ME2}(a), \ref {ME2}(b)).

\section { Conclusion an future works}

We have provided a new definition of privacy for the decentralized algorithms. We have proposed a new problem, the {\it decentralized exploration problem}, where players sampled from a distribution collaborate to identify a near-optimal arm with a fixed confidence, while ensuring privacy to players and minimizing the communication cost.  
We have designed and analyzed a generic algorithm for this problem: {\sc Decentralized Elimination} uses any best arm identification algorithm as an ArmSelection subroutine.
Thanks to the generality of the approach, we have extended the analysis of the algorithm to the case where the distributions of rewards are not stationary. 
Finally, our experiments suggest that {\it successive elimination} algorithms are better suited for the {\it decentralized exploration problem} than {\it explore-then-commit} algorithms.

Future work may focus on user-dependent best arms.
When Assumption 2 does not hold, {\sc Decentralized Elimination} finds with high probability the best arm of the most frequent players. 
However, in lot of applications the players can observe a context before choosing an arm. The extension of the proposed approach to {\it contextual bandits} is not straightforward because 
to collaborate for building a model, the players have to exchange messages about their favorite arms and their contextual variables, that also contain private information.

\onecolumn

\section {Proofs}

\paragraph {Theorem \ref {theorem:private}.}
{\it Using any $\operatorname{ArmSelection}$ subroutine,  {\sc Decentralized Elimination} is an $(\epsilon,\eta)$-private algorithm, that finds an $\epsilon$-optimal arm with a failure probability $\delta \leq \eta^{\lfloor \frac {\log \delta}{\log \eta} \rfloor}$ and that exchanges at most $\lfloor \frac {\log \delta}{\log \eta} \rfloor K -1$ messages.}

\begin {proof} 

The proof of Theorem \ref {theorem:private} is composed of three parts.
\paragraph{Part 1: $(\epsilon,\eta)$-privacy.}

Let $E_{l^n}=\{ \mathcal {K}^n(l^n) \cap \mathcal {K}_\epsilon = \emptyset \}$ be the event denoting that there is no $\epsilon$-optimal arm in the remaining set of arm $\mathcal {K}^n(l^n)$ at epoch $l^n$, and $\neg E_{l^n}$ be the event denoting that there is at least an $\epsilon$-optimal arm in the remaining set of arm $\mathcal {K}^n(l^n)$ at epoch $l^n$.

As {\sc Decentralized Exploration} ($\mathcal {A}$) performs an $\operatorname{ArmSelection}$ subroutine on each player, Property 1 ensures that for any player at epoch $l^n$:
\begin {align*}
& \mathds{P} \left( E_{l^n} | \mathcal{H}_{t^n-1}, \mathcal {A} , \neg E_{l^n-1} \right) 
\leq \eta \times f(l^n).
\end {align*}

For the sake of simplicity, in the following we will omit the dependence on $\mathcal {A}$ of probabilities.

The message $\lambda^n_k$ is sent by player $n$ as soon as the arm $k$ is eliminated from $\mathcal {K}^n(l^n)$ (see lines $17-18$ algorithm \ref {DSE}). Hence, we have:
\begin {align*} 
& \mathds{P} \left( E_{l^n} | \mathcal{M}_n , \neg E_{l^n-1} \right) 
 = \mathds{P} \left(E_{l^n} | \mathcal{H}_{t^n-1} , \neg E_{l^n-1}\right)
 \leq \eta \times f(l^n),
\end {align*}
where $t^n(l^n)$ is the time where epoch $l^n$ has begun.

To infer what arm is an $\epsilon$-optimal arm for player $n$ on the basis of $\mathcal{M}_n$ and $\mathcal {A}$, we first consider the favorable case for the adversary, where player $n$ has sent $K-1$ elimination messages which corresponds to epoch $l^n=L$. 
Using Property 1 of the subroutine used by $\mathcal {A}$ and the set of messages $\mathcal{M}_n$ the adversary can infer that:
\begin {align*} 
\mathds{P} \left( \{ \mathcal {K}^n(L) \not\subseteq \mathcal {K}_\epsilon\} | \neg E_{L-1} \right) 
& = \sum_{l^n=1}^L \mathds{P} \left( E_{l^n} | \mathcal{M}_n,\neg E_{l^n-1}  \right) \\
& \leq \eta \sum_{l^n=1}^L f(l^n) = \eta.
\end {align*}
The previous equality holds since if at epoch $l^n$ the event $\{ \mathcal {K}^n(l^n) \not\subseteq \mathcal {K}_\epsilon\}$ holds, then it holds also for all following epochs. Then the inequality is obtained by applying Property 1 to each element of the sum.
Hence, if $l^n=L$ knowing the set of messages $\mathcal {M}_n$ and Property 1, the adversary cannot infer what arm is an $\epsilon$-optimal arm for player $n$ with a probability higher that $1-\eta$.

Otherwise if $l^n < L$ then $\mathcal {K}^n(L) \subset \mathcal {K}^n(l^n)$, which implies that:
\begin {align*} 
& \mathds{P} \left( \{ \mathcal {K}^n(l^n) \not\subseteq \mathcal {K}_\epsilon\} | \mathcal{M}_n , \neg E_{l^n-1} \right)  \geq \mathds{P} \left( \{ \mathcal {K}^n(L) \not\subseteq \mathcal {K}_\epsilon\} | \mathcal{M}_n , \neg E_{L-1} \right).
\end {align*}

Hence, if $l^n < L$ the adversary cannot infer what arm is an $\epsilon$-optimal arm with a probability higher that $1-\eta$.

\paragraph{Part 2: Low probability of failure.}
An arm is eliminated when  the events $\{k \notin {\mathcal {K}}^n(l^n)\}$ occur for $\lfloor \frac {\log \delta}{\log \eta }\rfloor$ independent players. 
Assumption \ref {assump:active} ($\forall n$ $\in \mathcal {N}$, $P_x(x=n) \neq 0$) and Property 2 ensures that it exists a time $t=\sum_{n=1}^N t^n$ such that for $K-1$ arms, there are $\lfloor \frac {\log \delta}{\log \eta }\rfloor$ voting players.
Moreover, Property 1 implies that $\forall n \in \mathcal {N}$, $\forall l^n$:
\begin {displaymath}
\mathds{P} \left( \{ \mathcal {K}^n(l^n) \not\subseteq \mathcal {K}_\epsilon \} | \mathcal {M}_n,  \neg E_{l^n} \right) \leq \eta \times f(l^n).
\end {displaymath}

Hence, the $\lfloor \frac {\log \delta}{\log \eta }\rfloor$ independent voting players eliminate the $\epsilon$-optimal arm with a probability at most:
\begin {displaymath}
\mathds{P} \left( \{ \mathcal {K}(l) \not\subseteq \mathcal {K}_\epsilon \} | \mathcal {M}, \neg E_{l} \right) \leq \left(\eta \times f(l) \right)^{\lfloor \frac {\log \delta}{\log \eta }\rfloor},
\end {displaymath}

where $\mathcal {K}(l)$ denotes the shared set of remaining arms at elimination epoch $l$ (see line 7 of Algorithm \ref {DSE}), 
and $\mathcal {M}=\mathcal {M}_1 \cup \mathcal {M}_2 \cup ... \cup \mathcal {M}_N$.

If the algorithm fails, then the following event occurs : at stopping time, $\exists k \in \mathcal {K}(L), k \notin \mathcal{K}_{\epsilon}$. Using the union bound, we have:
\begin {align*}
\mathds{P} \left( \{ \mathcal {K}(L) \not\subseteq \mathcal {K}_\epsilon \} | \mathcal {M} , \neg E_{L-1} \right) & \leq \sum_{l=1}^{L} \left(\eta \times f(l) \right)^{\lfloor \frac {\log \delta}{\log \eta }\rfloor} \\
& \leq \eta^{\lfloor \frac {\log \delta}{\log \eta }\rfloor}.
\end {align*}

Finally notice that:
\begin {align*}
\eta^{\lceil \frac {\log \delta}{\log \eta }\rceil} \leq \delta = \eta^{\frac {\log \delta}{\log \eta }} \leq \eta^{\lfloor \frac {\log \delta}{\log \eta }\rfloor}.
\end {align*}

\paragraph {Part 3: Low communication cost.}
The index of each arm is sent to other players no more than once per player (see line $17$ of the algorithm \ref {DSE}).
When $\lfloor \frac {\log \delta}{\log \eta} \rfloor$ messages have been sent for an arm, this arm is eliminated for all players (see lines $4-9$ of the algorithm \ref {DSE}).

Thus $\lfloor \frac {\log \delta}{\log \eta} \rfloor (K-1)$ messages are sent to eliminate the suboptimal arms. Then, at most $\lfloor \frac {\log \delta}{\log \eta} \rfloor-1$ messages have been sent for the remaining arm. Thus, the number of sent messages is at most $\lfloor \frac {\log \delta}{\log \eta} \rfloor K -1$.

\end{proof}

\paragraph {Theorem \ref {theorem:sample}.}
{\it Using any $\operatorname{ArmSelection}(\epsilon,\eta,\mathcal{K})$ subroutine,  with a probability higher than \\$(1-\delta) \left(1-I_{1-p^*}\left(T_{P_{x,\bf y}}-T_{P_{\bf y}}, 1+ T_{P_{\bf y}}\right)\right)^{\lfloor \frac {\log \delta}{\log \eta} \rfloor}$ {\sc Decentralized Elimination} stops after at most:}

\begin {displaymath}
T_{P_{x,{\bf y}}} =
\frac{T_{P_{\bf y}}}{p^*}
+\frac {1}{(p^*)^2} \sqrt{\frac{p^*T_{P_{\bf y}}}{2}\log \frac {1}{\delta}} +
\frac{1}{2(p^*)^2} \log \frac {1}{\delta}
\text {\it samples in } P_{x,\bf y},
\end {displaymath}
{\it  where $I_a(b,c)$ denotes the incomplete beta function evaluated at $a$ with parameters $b,c$.}

\begin {proof}
 Let $T_{P_{x,\bf y}}$ be the number of samples in $P_{x,\bf y}$ when {\sc Decentralized Elimination} stops with high probability, and
$T_{P_{{\bf y}}}$ be the time step where the $\operatorname{ArmSelection}$ subroutine stops with high probability.
Let $T_n$ be the number of samples of player $n$ at time $T$. 
$T_n$ is a binomial law of parameters $T, P_x (x=n)$.
Hence, we have: 

\begin {displaymath}
\mathds{E}_{P_x} [T_n] =  P_x (x=n)T.
\end {displaymath}

Let $\mathcal {B}_{\delta,\eta}$ be the set of players that have the $\lfloor \frac {\log \delta}{\log \eta} \rfloor$ highest $T_n$.
The algorithm does not stop, if the following event occurs: $E_1=\{\exists n \in \mathcal {B}_{\delta,\eta}, T_n < T_{P_{\bf y}}\}$. 

Applying Hoeffding inequality, we have:
\begin {displaymath}
\mathds {P} \left(  T_n - P_x (x=n) T\leq -\epsilon \right) \leq
\exp (-\frac{2\epsilon^2}{T})
\end {displaymath}

When $\neg E_1$ occurs, $\forall n \in \mathcal {B}_{\delta,\eta}$ we have with a probability at most $\delta$:

\begin {displaymath}
T_{P_{\bf y}} - P_x (x=n) T \leq - \sqrt {\frac {T}{2} \log \frac {1}{\delta}}.
\end {displaymath}

\begin {displaymath}
\Leftrightarrow 
- P_x (x=n) T + \sqrt {\frac {T}{2} \log \frac {1}{\delta}} + T_{P_{\bf y}} \leq 0,
\end {displaymath}

\begin {displaymath}
\Leftrightarrow \sqrt{T} \geq \frac{1}{2P_x (x=n)} \left( \sqrt {\frac {1}{2} \log \frac {1}{\delta}} + \sqrt {\frac {1}{2} \log \frac {1}{\delta} + 4P_x (x=n)T_{P_{\bf y}}}\right), 
\end {displaymath}

\begin {displaymath}
\Leftrightarrow T \geq 
\frac{1}{4(P_x (x=n))^2} \left( \sqrt {\frac {1}{2} \log \frac {1}{\delta}} + \sqrt {\frac {1}{2} \log \frac {1}{\delta} + 4P_x (x=n)T_{P_{\bf y}}}\right)^2
\end {displaymath}

Then, when $\neg E_1$ occurs we have with a probability at most $\delta$:
\begin {displaymath}
T \geq  \frac{1}{4(p_{\delta,\eta})^2} \left( \sqrt {\frac {1}{2} \log \frac {1}{\delta}} + \sqrt {\frac {1}{2} \log \frac {1}{\delta} + 4p_{\delta,\eta}T_{P_{\bf y}}}\right)^2,
\end {displaymath}
where  $p_{\delta,\eta} = \min_{n \in \mathcal {B}_{\delta,\eta}}  P_x (x=n)$.

Hence, if $E_1$ does not occur, then we have with a probability at least $1-\delta$:
\begin {displaymath}
T < \frac{1}{4(p_{\delta,\eta})^2} \left( \sqrt {\frac {1}{2} \log \frac {1}{\delta}} + \sqrt {\frac {1}{2} \log \frac {1}{\delta} + 4p_{\delta,\eta}T_{P_{\bf y}}}\right)^2,
\end {displaymath}

\begin {displaymath}
\Leftrightarrow T < \frac{1}{4(p_{\delta,\eta})^2} \log \frac {1}{\delta}+ 
\frac{T_{P_{\bf y}}}{p_{\delta,\eta}} + \frac{1}{2(p_{\delta,\eta})^2}\sqrt {\frac {1}{2} \log \frac {1}{\delta}} \sqrt{{\frac {1}{2} \log \frac {1}{\delta}}+ 4p_{\delta,\eta}T_{P_{\bf y}}},
\end {displaymath}

\begin {displaymath}
\Rightarrow T < \frac{1}{4(p_{\delta,\eta})^2} \log \frac {1}{\delta}+ 
\frac{T_{P_{\bf y}}}{p_{\delta,\eta}} +
\frac {1}{4(p_{\delta,\eta})^2} \log \frac {1}{\delta}
+\frac {1}{(p_{\delta,\eta})^2} \sqrt{\frac{p_{\delta,\eta}T_{P_{\bf y}}}{2}\log \frac {1}{\delta}},
\end {displaymath}


\begin {displaymath}
\Rightarrow T < T_{P_{x,\bf y}}=
\frac{T_{P_{\bf y}}}{p_{\delta,\eta}}
+\frac {1}{(p_{\delta,\eta})^2} \sqrt{\frac{p_{\delta,\eta}T_{P_{\bf y}}}{2}\log \frac {1}{\delta}} +
\frac{1}{2(p_{\delta,\eta})^2} \log \frac {1}{\delta}.
\end {displaymath}

Let $\mathcal {N}_M$ bet the set of the $M=\lfloor \frac {\log \delta}{\log \eta} \rfloor$ most likely players.
Let $n^*=\arg \min_{n \in \mathcal {N}_M} P_x (x=n)
$, and $p^* = \min_{n \in \mathcal {N}_M} P_x (x=n)$.

Now, we consider the following event: $E_2=\{ n^* \notin \mathcal{B}_{\delta,\eta}\}$. By the definition of $\mathcal{B}_{\delta,\eta}$, the event $E_2$ is equivalent to the event $\{ T_{n^*} < T_{P_{\bf y}}\}$. Then, we have:
\begin {displaymath}
\mathds {P} \left( T_{n^*} < T_{P_{\bf y}}\right) = I_{1-p^*}\left(T_{P_{x,\bf y}}-T_{P_{\bf y}}, 1+ T_{P_{\bf y}}\right),
\end {displaymath}
where $I_a(b,c)$ denotes the incomplete beta function evaluated at $a$ with parameters $b,c$.

Finally, with a probability at least $(1-I_{1-p^*}\left(T_{P_{x,\bf y}}-T_{P_{\bf y}}, 1+ T_{P_{\bf y}}\right))^{\lfloor \frac {\log \delta}{\log \eta} \rfloor}$, we have $p_{\delta,\eta}=p^*$.

\end {proof}

\paragraph {\bf Corollary \ref {coro:median}}. {\it With a probability higher than $(1-\delta) \left(1-I_{1-p^*}\left(T_{P_{x,\bf y}}-T_{P_{\bf y}}, 1+ T_{P_{\bf y}}\right)\right)^{\lfloor \frac {\log \delta}{\log \eta} \rfloor}$ {\sc Decentralized Median Elimination}  stops after:}
\begin {displaymath}
\mathcal {O}\left(\left(   
\frac {K}{\lfloor \frac {\log \delta}{\log \eta} \rfloor \epsilon^2} 
+
\frac{1}{p^*}\right) \frac{1}{p^*} \log \frac {1}{\delta} \right)
 \text {\it samples in } P_{x,\bf y}.
\end {displaymath}

\begin {proof}

We have:
\begin {align*}
& \eta^{\lfloor \frac {\log \delta}{\log \eta} \rfloor} \leq \delta = \eta^{\frac {\log \delta}{\log \eta}} \leq \eta^{\lfloor \frac {\log \delta}{\log \eta} \rfloor} \\
& \Rightarrow \frac {1}{\delta} \geq \frac {1}{\eta^{\lfloor \frac {\log \delta}{\log \eta} \rfloor}} \\
& \Leftrightarrow \log \frac {1}{\eta} \leq \frac {1}{\lfloor \frac {\log \delta}{\log \eta} \rfloor}\log \frac {1}{\delta}
\end {align*}

{\sc Median Elimination} algorithm \cite {EMM2002} finds an $\epsilon$-optimal arm with a probability at least $1-\eta$ , and needs at most:

\begin {displaymath}
T_{P_{\bf y}}=\mathcal {O}\left(\frac {K}{\epsilon^2} \log \frac {1}{\eta}\right) \leq \mathcal {O}\left(\frac {K}{\lfloor \frac {\log \delta}{\log \eta} \rfloor\epsilon^2} \log \frac {1}{\delta} \right)\text { samples in } P_{\bf y}.
\end {displaymath}

Then the use of Theorem \ref {theorem:sample} finishes the proof.

\end {proof}

\paragraph {\bf Corollary \ref {coro:successive}.} {\it With a probability higher than $(1-\delta) \left(1-I_{1-p^*}\left(T_{P_{x,\bf y}}-T_{P_{\bf y}}, 1+ T_{P_{\bf y}}\right)\right)^{\lfloor \frac {\log \delta}{\log \eta} \rfloor}$  {\sc Decentralized Successive Elimination}  stops after:}
\begin {align*}
 \mathcal {O} \left( 
\left(   
\frac {K}{\lfloor \frac {\log \delta}{\log \eta} \rfloor \epsilon^2} 
+
\frac{1}{p^*}\right)\frac {1}{p^*}\log \frac {1}{\delta}
+\frac {K}{p^*\epsilon^2} \log K \sqrt {\log \frac {1}{\delta}}
\right)
\text {\it samples in } P_{x,\bf y}, 
\end {align*}
{\it samples in } $P_{x,\bf y}$.

\begin {proof}

{\sc Successive Elimination} algorithm \cite {EMM2002} finds an $\epsilon$-optimal arm with a probability at least $1-\eta$, and needs at most:
 
\begin {displaymath}
T_{P_{\bf y}} = \mathcal {O} \left( \frac {K}{\epsilon^2} \log \frac {K}{\eta} \right) 
\leq \mathcal {O} \left( \frac {K}{\epsilon^2} \left(
 \log K +
\frac {1}{\lfloor \frac {\log \delta}{\log \eta} \rfloor} \log \frac {1}{\delta} \right)\right)
\end {displaymath}
{samples in } $P_{x,\bf y}$.
Then using Theorem \ref {theorem:sample}, we have:
\begin {align}
T_{P_{x,{\bf y}}} & =  
\frac{T_{P_{\bf y}}}{p^*}
+\frac {1}{(p^*)^2} \sqrt{\frac{p^*T_{P_{\bf y}}}{2}\log \frac {1}{\delta}} +
\frac{1}{2(p^*)^2} \log \frac {1}{\delta}, \\
& \leq \mathcal {O} \left( \frac {K}{p^*\epsilon^2} \left(
 \log K +
\frac {1}{\lfloor \frac {\log \delta}{\log \eta} \rfloor} \log \frac {1}{\delta} \right) +\frac {1}{(p^*)^2} \sqrt{ \frac {Kp^*}{\epsilon^2} \left(
 \log K +
\frac {1}{\lfloor \frac {\log \delta}{\log \eta} \rfloor} \log \frac {1}{\delta} \right)\log \frac {1}{\delta}} +
\frac{1}{(p^*)^2} \log \frac {1}{\delta}
\right),\\
& \leq
\mathcal {O} \left( 
\frac {K}{p^*\lfloor \frac {\log \delta}{\log \eta} \rfloor\epsilon^2} \log \frac {1}{\delta} 
+
\frac{1}{(p^*)^2} \log \frac {1}{\delta}
+\frac {K}{p^*\epsilon^2} \log K \sqrt {\log \frac {1}{\delta}}
\right), \\
& \leq \mathcal {O} \left( 
\left(   
\frac {K}{\lfloor \frac {\log \delta}{\log \eta} \rfloor \epsilon^2} 
+
\frac{1}{p^*}\right)\frac {1}{p^*}\log \frac {1}{\delta}
+\frac {K}{p^*\epsilon^2} \log K \sqrt {\log \frac {1}{\delta}}
\right).
\end {align}
\end {proof}

\paragraph {\bf Theorem \ref {th:DSER3}.} 
{\it For  $K\geq2$,  $\delta \in (0,0.5]$, $\epsilon\geq \delta / K$,
for the sequences of rewards where Assumption 4 holds, DSER3 is an $(\epsilon,\eta)$-private algorithm, that exchanges at most $\lfloor \frac {\log \delta}{\log \eta} \rfloor K -1$ messages, that finds an $\epsilon$-optimal arm with a probability at least $(1-\delta) \left(1-I_{1-p^*}\left(T_{P_{x,\bf y}}-T_{P_{\bf y}}, 1+ T_{P_{\bf y}}\right)\right)^{\lfloor \frac {\log \delta}{\log \eta} \rfloor}$, and that stops after: }
\begin {align*}
\mathcal {O} \left( 
\left(   
\frac {K}{\lfloor \frac {\log \delta}{\log \eta} \rfloor \epsilon^2} 
+
\frac{1}{p^*}\right)\frac {1}{p^*}\log \frac {1}{\delta}
+\frac {K}{p^*\epsilon^2} \log K \sqrt {\log \frac {1}{ \delta}}
\right)
\end {align*}
{\it samples in } $P_{x,\bf y}$.

\begin {proof}

Theorem \ref {th:DSER3} is a straightforward application of Theorem \ref {theorem:sample}, where $T_{P_{\bf y}}$ is stated in Theorem 1 \cite {AFM2017}.
\end {proof}

\paragraph {\bf Theorem \ref {th:DSER4}.} 
{\it For  $K\geq2$,  $\epsilon \geq \frac {\eta}{K}$, $\varphi \in (0,1]$, $\epsilon \geq \delta/K$ for any sequences of rewards that can be splitted into sequences where Assumption 4 holds, {\sc DSER4} is an $(\epsilon,\eta)$-private algorithm, that exchanges on average at most $\varphi T(\lfloor \frac {\log \delta}{\log \eta} \rfloor K -1)$ messages, and that plays, with an expected probability at most $\delta+\varphi T I_{1-p^*}\left(T_{P_{x,\bf y}}-T_{P_{\bf y}}, 1+ T_{P_{\bf y}}\right)^{\lfloor \frac {\log \delta}{\log \eta} \rfloor}$, a suboptimal arm on average no more than: }
\begin {align*}
  \mathcal {O} \left( \frac{1}{\epsilon^2} \sqrt{\frac{\left(   
\frac {K}{p^*} \log K +
\frac {K}{\lfloor \frac {\log \delta}{\log \eta} \rfloor } 
\right)\frac {S}{p^*}\log \frac {1}{\delta}}{\delta} }
 +\frac {1}{(p^*)^2}\log \frac {1}{\delta}
\right)
\end {align*}
{\it times, where $S$ is the number of switches of best arms, $\varphi$ is the probability of reset in {\sc SER4}, $T$ is the time horizon, and the expected values are taken with respect to the randomization of resets.}

\begin {proof}

The upper bound of the expected number of times a suboptimal arm is played by {\sc SER4}, is stated in Corollary 2 \cite {AFM2017}. 
Then this upper bound is used in Theorem \ref {theorem:sample} to state the upper bound of the expected number of times a suboptimal arm is played using {\sc DSER4}:
\begin {align}
T_{P_{x,{\bf y}}} & \leq \mathcal {O} \left( \frac{\varphi}{\delta} 
\left(   
\frac {K}{\lfloor \frac {\log \delta}{\log \eta} \rfloor \epsilon^2} 
+
\frac{1}{p^*}\right)\frac {1}{p^*}\log \frac {1}{\delta}
+ \frac{\varphi}{\delta}  \frac {K}{p^*\epsilon^2} \log K \sqrt { \log \frac {1}{\delta}} + \frac{S}{\varphi} 
\right), \\
& \leq \mathcal {O} \left( \frac{\varphi}{\delta}
\left(   
\frac {K}{p^*\epsilon^2} \log K +
\frac {K}{\lfloor \frac {\log \delta}{\log \eta} \rfloor \epsilon^2} 
+
\frac{1}{p^*}\right)\frac {1}{p^*}\log \frac {1}{\delta}
+\frac{S}{\varphi} 
\right), \\
& \leq \mathcal {O} \left( \frac{\varphi}{\delta} 
\left(   
\frac {K}{p^*\epsilon^2} \log K +
\frac {K}{\lfloor \frac {\log \delta}{\log \eta} \rfloor \epsilon^2} 
\right)\frac {1}{p^*}\log \frac {1}{\delta}
+\frac{S}{\varphi} +\frac {1}{(p^*)^2}\log \frac {1}{\delta}
\right).
\end {align}

Then setting: $$\varphi = \sqrt{\frac{S\delta}{\left(   
\frac {K}{p^*} \log K +
\frac {K}{\lfloor \frac {\log \delta}{\log \eta} \rfloor } 
\right)\frac {1}{p^*}\log \frac {1}{\delta}} },$$
we obtain:
\begin {align}
T_{P_{x,{\bf y}}} \leq  \mathcal {O} \left( \frac{1}{\epsilon^2} \sqrt{\frac{\left(   
\frac {K}{p^*} \log K +
\frac {K}{\lfloor \frac {\log \delta}{\log \eta} \rfloor } 
\right)\frac {S}{p^*}\log \frac {1}{\delta}}{\delta} }
 +\frac {1}{(p^*)^2}\log \frac {1}{\delta}
\right).
\end {align}

The expected number of resets is $\varphi T$. Theorem \ref {theorem:sample} provides the success probability of each run of {\sc Decentralized Elimination}, which states the expected failure probability of {\sc DSER4}. Then using Theorem \ref {theorem:private} the expected upper bound of the number of exchanged messages is stated. 
\end {proof}

\end {document}